\documentclass[letterpaper, 10 pt, conference]{ieeeconf}  

\IEEEoverridecommandlockouts                              

\let\labelindent\relax

\usepackage[
    style=ieee,
    doi=false,
    isbn=false,
    url=false,
    eprint=false,
    backend=bibtex,
    natbib=true,
    minnames=1,
    maxcitenames=1
    ]{biblatex}

\usepackage{hyperref}
\hypersetup{
    colorlinks=true,
    linkcolor=black,
    urlcolor=blue,
    citecolor=black,
    }
    
\usepackage[shortcuts]{extdash}
\usepackage{graphicx}
\usepackage{mathtools}
\usepackage{dcolumn}
\usepackage{gensymb}
\usepackage{perl_acronyms}
\usepackage{xcolor}
\usepackage{subcaption}
\usepackage{multirow}
\usepackage{amsfonts}
\usepackage{booktabs}
\usepackage{arydshln}
\usepackage{siunitx}
\usepackage{todonotes}
\usepackage{algorithm}
\usepackage{algpseudocode}
\usepackage[inline]{enumitem}
\usepackage{array}
\usepackage{flushend}
\newcolumntype{P}[1]{>{\centering\arraybackslash}p{#1}}

\title{Sonar-MASt3R: Real-Time Opti-Acoustic Fusion in Turbid, Unstructured Environments}

\author{Amy Phung$^{1,2}$, Richard Camilli$^{2}$
\thanks{This work was supported by the Strategic Environmental Research and Development Program Grant W912HQ24P0024. Amy Phung would like to acknowledge financial support from the National Science Foundation Graduate Research Fellowship (No. 2141064), from the National Aeronautics and Space Administration (NASA) through the FINESST program (No. 80NSSC23K1391)}
\thanks{$^{1}$Applied Ocean Physics and Engineering, Woods Hole Oceanographic Institution, Deep Submergence Laboratory, Woods Hole, MA, USA}
\thanks{$^{2}$Massachusetts Institute of Technology, Cambridge, MA, USA
         {\tt\footnotesize aphung@mit.edu}}%
\thanks{Supplemental Video: \url{https://youtu.be/LkW0TpIiwBA}}
\thanks{The code, dataset, and visualization tools used to generate the figures in this paper are available online at \url{https://sonar-mast3r.github.io/}}
}

\begin{document}

\maketitle
\begin{abstract}
\label{sec:abstract}
Underwater intervention is an important capability in several marine domains, with numerous industrial, scientific, and defense applications. However, existing perception systems used during intervention operations rely on data from optical cameras, which limits capabilities in poor visibility or lighting conditions. Prior work has examined opti-acoustic fusion methods, which 
use sonar data to resolve the depth ambiguity of the camera data while using camera data to resolve the elevation angle ambiguity of the sonar data. 
However, existing methods cannot achieve dense 3D reconstructions in real-time, and few studies have reported results from applying these methods in a turbid environment.
In this work, we propose the opti-acoustic fusion method Sonar-MASt3R, which uses MASt3R to extract dense correspondences from optical camera data in real-time and pairs it with geometric cues from an acoustic 3D reconstruction to ensure robustness in turbid conditions. Experimental results using data recorded from an ``opti-acoustic eye-in-hand'' configuration across turbidity values ranging from $<$0.5 to $>$12 NTU highlight this method's improved robustness to turbidity relative to baseline methods.

\end{abstract}
\section{Introduction}
\label{sec:introduction}
Underwater perception in turbid conditions remains a challenge for subsea intervention operations, which are important for a variety of tasks ranging from scientific exploration~\cite{billings2022hybrid} to infrastructure construction~\cite{rizzini2017integration} and maintenance~\cite{zhang2024advanced}. Today, intervention operations predominantly rely on optical cameras for real-time feedback~\cite{cai2023autonomous, cieslak2015autonomous, simetti2020autonomous}, which provide operators with sufficient information to assess obstacles and manipulation targets in the environment. However, optical cameras lack robustness to poor lighting or visibility conditions, which can slow or even halt operations. 

Perception during intervention operations poses particular challenges because tasks involving disturbing soft sediments, scraping biofouling, drilling, or excavation can quickly increase turbidity. To support effective manipulation in these conditions, perception methods must satisfy two key requirements: (i) operate at real-time rates to provide operators with continuous visual feedback, and (ii) provide sufficiently accurate geometric and color correspondences to enable reliable object identification.

Prior research has proposed various methods for opti-acoustic fusion, which leverage the complementary strengths of optical and acoustic sensors: optical cameras provide high-resolution color, texture, and semantic information, while acoustic sensors deliver robust spatial geometry and depth information, and remain functional in degraded visibility conditions~\cite{singh2024opti, qadri2024aoneus, babaee20133}. However, existing methods cannot achieve dense 3D reconstructions in real-time, and few studies have reported results from applying these methods in a turbid environment.

In this work, we present Sonar-MASt3R, an ``opti-acoustic eye-in-hand'' perception method designed for robust perception in turbid underwater environments. Sonar-MASt3R uses MASt3R to extract dense correspondences from optical camera streams in real-time, and fuses them with geometric information obtained from an acoustic 3D reconstruction to ensure the reconstruction maintains absolute length scale. This method enables us to use the high-resolution visual and geometric cues of cameras for dense reconstruction when sufficient visual structure is available, while maintaining robustness to visibility degradation with the sonar data. In summary, the key contributions of this work are as follows:
\begin{itemize}
    \item \textbf{Dense metric-scale 3D reconstruction in real-time} Sonar-MASt3R fuses dense optical correspondences from MASt3R by using acoustic data to apply an absolute scale correction to the unscaled MASt3R output.
    \item \textbf{Adapting the use of optical and acoustic data based on visibility conditions} The method leverages dense optical features when visibility allows, while maintaining robustness by reverting to the acoustic reconstruction in degraded conditions.
    \item \textbf{Experimental validation in turbid conditions} We present qualitative and quantitative results from using the method in turbidity levels ranging from $<$0.5 to 12 NTU.
    \item \textbf{New opti-acoustic dataset} We release a new opti-acoustic dataset recorded in a test tank with calibrated turbidity levels and a cluttered workspace.
\end{itemize}

\section{Related Work}
\label{sec:related}
Prior research has developed a variety of opti-acoustic perception methods for different applications. Early approaches primarily relied on geometric formulations, which leverage the epipolar geometry of the sensors to compute the 3D position of matching correspondences~\cite{negahdaripour2009opti, negahdaripour2007epipolar}. While these approaches have been widely used for extrinsic calibration with known targets~\cite{negahdaripour2009opti, chemisky2021portable, hurtos2010integration}, the need for feature correspondences makes them difficult to use with unknown workspaces, and results in a very sparse reconstruction. A contour-based approach proposed in~\cite{babaee20153, babaee2015improved} improves the robustness and reconstruction density of feature-based methods by fitting a contour to correspondences, but can only reconstruct discrete objects rather than the overall scene geometry and still provides insufficient resolution for object recognition. 

The AoNeuS method focused on accurate high-resolution 3D surfaces from opti-acoustic measurements captured over heavily-restricted baselines, where acquiring full 360 degree views of objects may not be possible~\cite{qadri2024aoneus}. Rather than focusing on correspondences, this method uses the raw optical and acoustic data to compute the optimal geometry by using a neural surface reconstruction method. 
While the AoNeuS method produces a high-resolution reconstruction, it requires significant computational resources to process and can take hours to optimize, with typical rendering times of $\sim$ 5 minutes per frame depending on resolution and dataset size~\cite{qadri2024aoneus}. 

To provide sufficient resolution for object recognition in real time, the OASIS method 
proposed a volumetric approach for reconstructing the workspace geometry using a wrist-mounted sonar, then projected data from a wrist-mounted camera on the reconstruction to aid in object recognition~\cite{oasis}. Although this method achieved sufficient resolution for object identification in real time, its sole reliance on acoustic data for reconstructing workspace geometry limited its ability to reconstruct small objects.

Other than the contour method presented in~\cite{babaee2015improved}, it is also worth noting that the aforementioned works do not evaluate the performance of these methods in turbid water. Recently, \cite{collado2025opti} introduced a real-time opti-acoustic 3D reconstruction method, which identifies regions of interest in camera images and uses corresponding sonar measurements to estimate range. Although the presented seawall and piling reconstruction results are compelling, the authors acknowledge that the method’s applicability to more complex scenes remains limited.



Computing dense 3D reconstructions in complex workspaces remains an open challenge for existing opti-acoustic methods. For optical datasets, MASt3R-SLAM proposed a keyframe-based approach which achieved dense 3D reconstruction in complex environments~\cite{mast3r-slam}. The method uses the reconstruction model MASt3R~\cite{mast3r} to obtain dense correspondences, which are then used to estimate the camera pose and align sequential frames. However, because MASt3R was trained on in-air datasets, MASt3R-SLAM performs poorly under turbid conditions when it selects a keyframe with no coherent features. Additionally, as a monocular system, it lacks an inherent sense of spatial scale and cannot incorporate external pose estimates provided in real-time. Consequently, its reconstructions are sometimes distorted since the data is aligned on a frame-by-frame basis.


\section{Method}
\label{sec:method}

\begin{figure*}[t]
  \centering
  \includegraphics[width=0.8\textwidth]{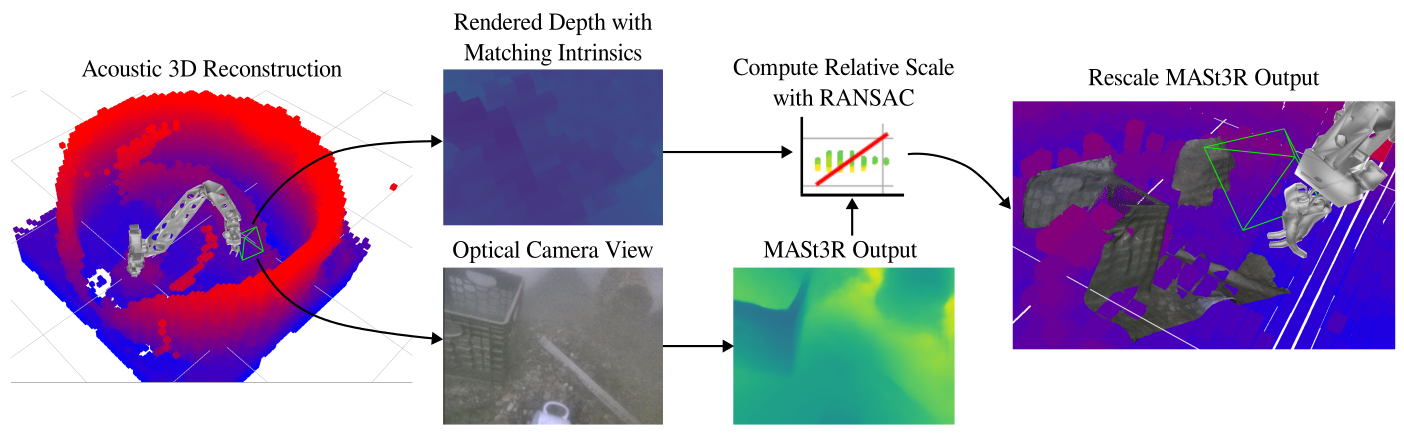}
  \caption{The optical camera intrinsics and extrinsics are used to render a depth image from the acoustic 3D reconstruction, which is used to correct the scale of the pointmap computed by MASt3R}
  \label{fig:workflow}
\end{figure*}

Sonar-MASt3R is a real-time, opti-acoustic 3D reconstruction method that fuses the dense optical features and correspondences from MASt3R with the geometry and scale information inferred from range-based acoustic data. We use an ``opti-acoustic eye-in-hand'' configuration similar to OASIS~\cite{oasis}, and a keyframe-based method built on MASt3R-SLAM~\cite{mast3r-slam} to process the incoming data in real time. 
The wrist-mounted optical camera and imaging sonar leverages the manipulator's dexterity to record an opti-acoustic dataset spanning the manipulator's reachable workspace. The manipulator's joint angle sensors and forward kinematics model provides pose information. The overall workflow is summarized in Figure~\ref{fig:workflow}.

\subsection{Acoustic 3D Reconstruction}
The process is initiated using the OASIS method described in~\cite{oasis} to compute an acoustic-only 3D reconstruction. This method uses a low-profile ``sweep'' trajectory that moves the arm minimally from its initial stowed position, making it safe to use in an unmapped environment. It is optimized for recording intersecting regions of the sonar data to quickly resolve the sonar's elevation angle ambiguity, and tilts the manipulator's wrist to maximize the use of the sonar's wide field-of-view (FOV). The output of OASIS is a voxel grid, which can be written as
\begin{align}
V(i_x, i_y, i_z) \to 
\begin{cases} 
    1 \text{ if occupied} \\
    0 \text{ otherwise}
\end{cases}
\end{align}
where $i_x, i_y, i_z$ are the corresponding indices for the voxel grid given an x,y,z position in the world frame. For this implementation, a voxel resolution of 5 cm is used.

Notably, using an acoustic-only method allows the trajectory to be executed at a much higher speed than would be optimal for an opti-acoustic dataset.
We do not use camera data recorded using this trajectory since the manipulator's speed results in a significant amount of motion blur. By design, this trajectory maintains a large standoff distance between the sensors and the workspace. While this has a marginal impact on the sonar's performance, the optical image quality rapidly degrades with range, particularly in turbid conditions.

\subsection{Real-time Opti-Acoustic Processing}
Once the acoustic 3D reconstruction has been computed, the incoming camera data is fused with the voxel grid data using a keyframe-based approach. For each new camera frame, the reconstruction process can be described as follows:
\begin{enumerate}
    \item Compare current frame with previous keyframe using MASt3R
    \item Use acoustic 3D reconstruction to rescale the MASt3R pointmap output
    \item Match current frame pointmap scale to keyframe
    \item Determine whether to add current frame as a new keyframe
    \item After adding new keyframes, run global optimization  
\end{enumerate} 

These steps will be detailed in the following subsections.

\subsection{Pair-wise Frame Processing}
\label{sec:pair-process}
The basis of this method relies on pair-wise image processing, consisting of two steps: MASt3R prediction, which generates dense pointmaps, features, and confidences from a pair of optical images, and a sonar-based rescaling step, which corrects the scale of the MASt3R outputs.

Before processing images with MASt3R, we rescale our camera's native 1600x1200 px resolution to 512x384 px since the maximum dimension of input images in MASt3R is 512.
Although MASt3R is typically used with a pair of unrectified images, we also rectify the fisheye camera's images before processing them. Since MASt3R's training dataset consists of in-air images from regular pinhole cameras, which have less radial distortion than fisheye cameras, using unrectified fisheye images with MASt3R results in distorted pointmaps. However, fisheye cameras are commonly used in underwater applications due to their wider field of view -- when a regular pinhole camera is placed inside a flat-port housing, the refractive index difference between the water and air inside the housing results in a significant decrease in the cameras's FOV. 

\textbf{MASt3R Prediction}: MASt3R takes in a pair of images $\mathcal{I}^i, \mathcal{I}^j \in \mathbb{R}^{H \times W \times 3}$, and outputs pointmaps $X \in \mathbb{R}^{H \times W \times 3}$, pointmap confidences $C \in \mathbb{R}^{H \times W \times 1}$, $d$-dimensional features $D \in \mathbb{R}^{H \times W \times d}$, and feature confidences $Q \in \mathbb{R}^{H \times W \times 1}$ for both images. The forward pass of MASt3R can be written as:
\begin{align}
    \mathcal{F}_M(\mathcal{I}^i, \mathcal{I}^j) \to X_i^i, X_i^j, C^i, C^j, D^i, D^j, Q^i, Q^j
\end{align}

Note that the pointmap outputs for image $i$ ($X_i^i$) and $j$ ($X_i^j$) are both specified with respect to image $i$, hence the matching subscript. 

\textbf{Rescale MASt3R Output with Acoustic Reconstruction:}
While MASt3R can extract 3D structure information from image pairs, the scale of the output pointmap is arbitrary and can vary between different pairs of images. To achieve metric scale, we use the acoustic 3D reconstruction to correct the MASt3R pointmap scale. 

Since the MASt3R pointmap $X_i^i$ consists of 3D positions of each pixel relative to the camera center, the optical depth image $D_O \in \mathbb{R}^{H \times W \times 1}$ can be computed using the norm of this pointmap 
\begin{align}
    D_O = \lVert X_i^i \rVert
\end{align}
Using the optical camera's intrinsics $K_i$ (after downsampling to the maximum MASt3R resolution) and pose $T_{Wf}$ (derived from the manipulator's joint angle sensors and forward kinematics model), a corresponding depth image is rendered from the acoustic reconstruction voxel grid $V$. This process, denoted as $\mathcal{F}_D(K_i, T_{Wf}, V)$ entails sampling along rays defined by the camera's projection model at intervals specified by the grid's voxel size for each pixel in the image. The corresponding acoustic depth image $D_A \in \mathbb{R}^{H \times W \times 1}$ can be denoted as 
\begin{align}
    D_A = \mathcal{F}_D(K_i, T_{Wf}, V)
\end{align}
The computed depth images $D_O$ and $D_A$ are illustrated in Figure~\ref{fig:workflow}.

The relative scale $s$ between $D_O$ and $D_A$ is estimated using RANSAC~\cite{fischler1981random}, which is a robust estimation technique with moderate tolerance to outliers. Prior to estimation, we construct subsets $D_O^{\prime} \subset D_O$ and $D_A^{\prime} \subset D_A$, 
which are defined by

\begin{align}
\Omega_D' = \left\{ (u,v) \in \Omega_D \;\middle|\;
\begin{aligned}
& D_O(u,v) > 0, \\
& D_A(u,v) > 0, \\
& D_O(u,v), D_A(u,v) \in \mathbb{R}, \\
& C_{ii}(u,v) > \mathrm{mean}(C_{ii})
\end{aligned}
\right\}
\end{align}

This subset contains the set of pixels where the optical and acoustic depth values are both positive and finite, and correspond to pixel values where the corresponding MASt3R pointmap confidence is within the upper half of the confidence distribution. Since the optical depth values are inferred from the MASt3R pointmap, this filtering ensures the points used for estimating scale contain a sufficient number of points for sampling while discarding lower-confidence depth values.





The computed scaling factor $s_\text{m}$ is applied to the original pointmap $X_i^i$ to construct the metric-scale pointmap $\bar{X}_i^i$
\begin{align}
    \bar{X}_i^i = s_\text{m} X_i^i
\end{align}

\subsection{Initialization}

During initialization, each incoming frame is used as both $\mathcal{I}^i$ and $\mathcal{I}^j$ in the pair-wise frame process described in Section~\ref{sec:pair-process}. Although there is no baseline difference when using the same image as the pair, MASt3R will still compute an unscaled pointmap $X_i^i$ and corresponding confidence values $C_i^i$. We select the first frame satisfying $\max(C_i^i) > \tau_{i}$, where $\tau_{i}$ is a predefined initialization confidence threshold, and designate it as the initial keyframe $k$. The metric-scale pointmap for the keyframe $\bar{X}_k^k$ is computed using the corresponding acoustic depth image. 



\subsection{Pointmap-Based Scale Refinement}
After initialization, each incoming frame $f$ is processed with the last keyframe $k$ using the process described in Section~\ref{sec:pair-process}. Although the corresponding acoustic depth image for frame $f$ can be used to compute its metric-scale pointmap $\bar{X}_f^f$, using this scale estimation alone produces inconsistent results (i.e., multiple views of an object may not appear aligned or consistently scaled). Therefore, to further refine the scale after the acoustic rescaling step, we compute a least squares solution between a subset containing valid matches from each of the pointmaps $\bar{X}_f^{f\prime} \subset \bar{X}_f^f$ and $\bar{X}_k^{k\prime} \subset \bar{X}_k^k$. These subsets contain the overlapping components of the pointmaps where the corresponding pointmap and feature confidences are above their respective thresholds $\tau_c, \tau_q$, and are defined by
\begin{align}
\label{eq:m_subset}
\Omega_M' = \left\{ 
\left\langle 
\begin{aligned}
(u_f,v_f) \\
(u_k,v_k) 
\end{aligned}
\right\rangle \in M \;\middle|\;
\begin{aligned}
& C_k(u_k,v_k) > \tau_c, \\
& C_f(u_f,v_f) > \tau_c, \\
& Q > \tau_q
\end{aligned}
\right\}
\end{align}
where
\[
Q = \sqrt{Q_f(u_f,v_f) \cdot Q_k(u_k,v_k)} 
\]

The set of matches $M$, which contains pixel mappings between the current frame and the last keyframe $(u_f, v_f) \mapsto (u_k, v_k)$ are identified using the projective data association step as described in~\cite{mast3r-slam}. This pointmap-based scale refinement $s_\text{p}$ is computed by solving
\begin{align}
\min_{s_\text{p}} \; \| \bar{X}_k^{k\prime} - s_\text{p} \cdot T_{kf} \bar{X}_f^{f\prime} \|^2
\end{align}

where $T_{kf}$ is the matrix that describes the transformation between frames $k$ and $f$. This matrix is derived from the manipulator pose data recorded with each frame.








\subsection{Keyframe Selection and Global Optimization}
Following the keyframe selection method in~\cite{mast3r-slam}, a new keyframe is introduced when the fraction of valid matches $\alpha_\text{match}$ or unique keyframe pixels $\alpha_\text{unique}$ falls below a threshold $\tau_k$. These fractions are defined as
\begin{align}
\label{eq:a_match}
\alpha_\text{match} =& \frac{\left| \Omega_M'\right|}{H*W} \\
\alpha_\text{unique} =& \frac{\left|\text{unique}(M_k)\right|}{H*W}
\end{align}
where $H$ and $W$ are the dimensions of the image, $\left| \Omega_M'\right|$ is the number of matches in the subset defined by $\Omega_M'$, and $\left|\text{unique}(M_k)\right|$ is the number of unique output pixels in the keyframe from the set of matches $M$. A new keyframe is added when 
\begin{align}
\text{min}(\alpha_\text{match}, \alpha_\text{unique}) < \tau_k
\end{align}

When each new keyframe is added, we run a global optimization process that improves the consistency of the overall map. All of the keyframes are stored in a factor graph, with edges added between each pair of keyframes if the match fraction between them (as defined by Equations~\ref{eq:m_subset} and~\ref{eq:a_match}) is above a threshold $\tau_f$. 

Our factor graph implementation is based on the approach described in~\cite{mast3r-slam}, with a modification to support separate optimization for each disconnected subgraph. Since MASt3R-SLAM requires feature matches between sequential frames for tracking, their implementation assumes that the factor graph always remains fully connected. However, in our case, there may not be feature matches between clusters of keyframes, especially in low-visibility conditions. 

The factor graph optimization process in~\cite{mast3r-slam} aligns the scale and poses of all keyframes to the first keyframe, which works for unscaled reconstructions with an arbitrary coordinate reference. However, if left unrectified in our implementation, the first frame's pose and scale errors can misalign the entire map. To improve the alignment, we solve for the optimal transformation matrix $X$ such that 
\begin{align}
\mathbf{T_{Wk}} \approx X \mathbf{\tilde{T}_{Wk}}
\end{align}
where $\mathbf{T_{Wk}}\in \mathbb{R}^{N\times4\times4}$ is a tensor containing all of the keyframe poses and $\mathbf{\tilde{T}_{Wk}}$ are the optimized poses computed from the factor graph optimization process. We apply this transformation matrix $X$ to all of the keyframe pointmaps before rendering the updated reconstruction.

\subsection{Recovery mode}
If the match fraction $\alpha_\text{match}$ falls below the threshold $\tau_r$ (which is smaller than $\tau_k$), we switch to a ``recovery mode,'' which typically occurs when the visibility drops below a critical threshold and there are insufficient features for matching the current frame to the last keyframe. In this mode, we keep a list of the 10 most recent frames and their average pointmap confidence values in a list, which is initialized starting with the last keyframe. During recovery, instead of comparing each new frame to the last keyframe, we compare it to the frame with the highest confidence in our list. This step prevents the method from stalling while looking for matches to the last available keyframe, which can happen when objects fade out of view due to low visibility.
\subsection{Color Correction}
\begin{figure}[h]
  \centering
\includegraphics[width=0.9\linewidth]{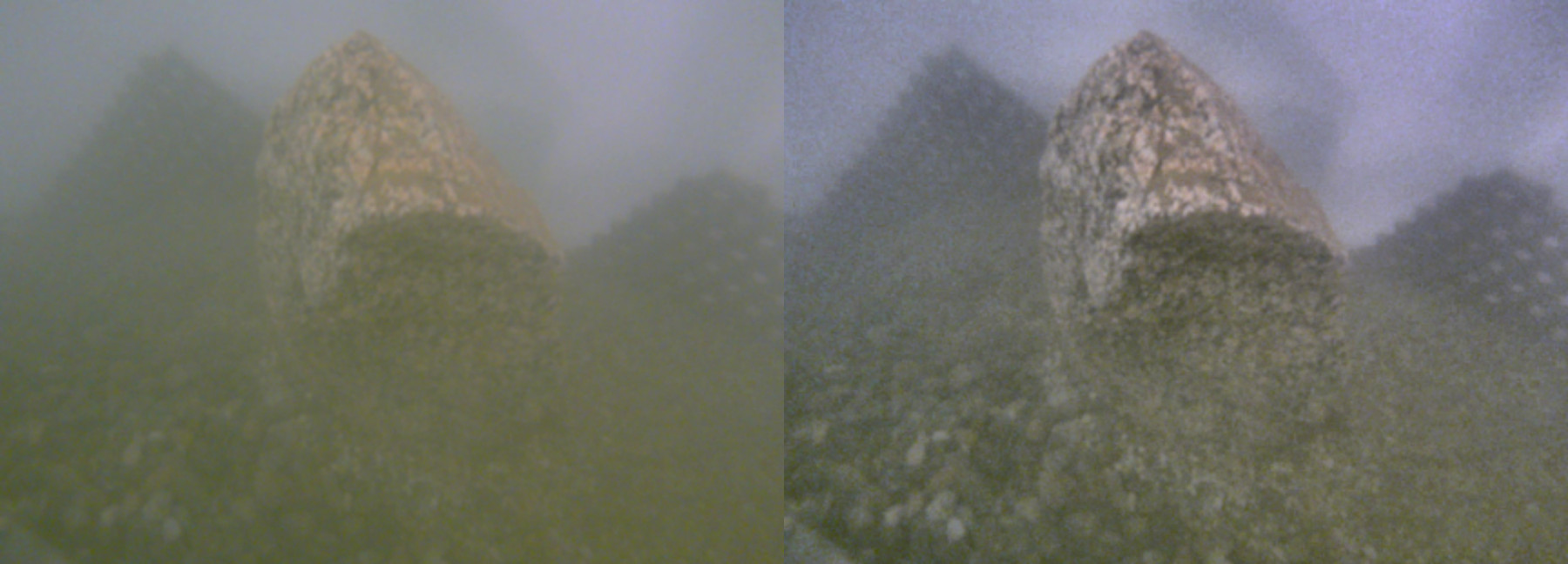}
  \caption{(left) Before and (right) after applying histogram equalization to an optical image of a granite boulder in $\sim$8 NTU}
  \label{fig:equalization}
\end{figure}

We applied a histogram-based correction before rendering the pointmaps to improve the color and contrast. Surprisingly, we found that MASt3R performed worse on color-corrected images, and thus we only applied this correction to the pointmaps after processing the original images with MASt3R. Results from applying this correction to an image are illustrated in~\ref{fig:equalization}.

\section{Dataset}
\label{sec:dataset}
For evaluation, we recorded a collection of tank-based datasets with turbidity values ranging from $<$0.5 to $>$12 NTU. At each NTU value, we recorded optical and acoustic data using both a ``sweep'' trajectory (used for the acoustic reconstruction), and an object-centric trajectory, which starts from the manipulator's stowed position and sequentially moves towards each of the objects in the tank.

\begin{figure}[h]
  \centering
\includegraphics[width=0.5\linewidth]{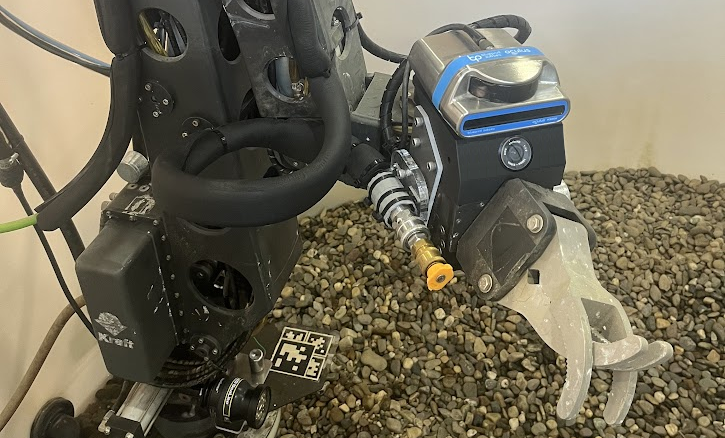}
  \caption{An ``opti-acoustic eye-in-hand'' configuration is used to record the datasets. The sonar is mounted level relative to the manipulator's wrist, and the camera is mounted 30 degrees off of vertical.}
  \label{fig:mount}
\end{figure}
\vspace{-8pt}

\begin{figure}[h]
  \centering
\includegraphics[width=0.8\linewidth]{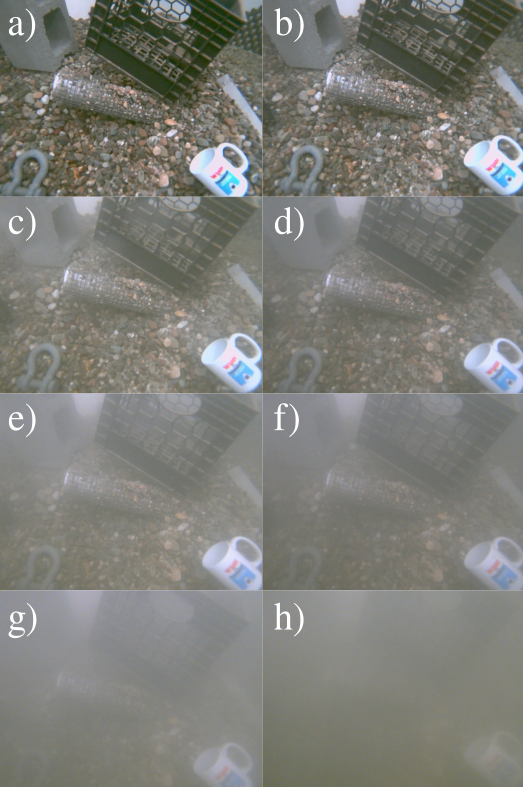}
  \caption{Illustration of different turbidity levels within each of the recorded datasets. NTU values: (a) $<0.5$, (b) 0.83, (c) 2.61, (d) 3.92, (e) 5.41, (f) 7.84, (g) 11.31, (h) 12.39}
  \label{fig:turbid-vis}
\end{figure}

These datasets were recorded in a 2.1 m diameter, 1.5 m deep tank, with a collection of objects with various materials and geometries. For sensing, an Oculus MT3000d multibeam imaging sonar (Blueprint Subsea) operating at 1.2MHz and a stellarHD Global Shutter Frame-Sync Subsea ROV/AUV USB Machine Vision Camera (DeepWater Exploration) were mounted on the wrist of a fixed-base seven degree-of freedom hydraulic manipulator arm (Kraft TeleRobotics). The 512-beam multibeam sonar has a horizontal FOV of 130$^\circ$ and a vertical FOV of 20$^\circ$, and recorded data at 10 FPS. The sonar was configured with a 2 m maximum range and a gain of 10.
The camera employs a 1/2.9-inch Omnivision OmniPixel 3-GS CMOS sensor with a maximum resolution of 1600 × 1200 and was operated at 10 FPS.

\begin{table}[h]
\centering 
\caption{Turbidity Measurements for Each Dataset}
\begin{tabular}{c|ccc|c}
\multicolumn{1}{l|}{Dataset} & \multicolumn{3}{l|}{Turbidity Measurements (NTU)} & \multicolumn{1}{l}{Average} \\ \hline
A & - & - & - & \textless{}0.5 \\
B & 1 & 0.8 & 0.7 & 0.83 \\
C & 2.54 & 2.63 & 2.65 & 2.61 \\
D & 3.8 & 3.87 & 4.08 & 3.92 \\
E & 5.15 & 5.48 & 5.6 & 5.41 \\
F & 8.13 & 7.57 & 7.82 & 7.84 \\
G & 10.76 & 11.75 & 11.42 & 11.31 \\
H & 12.37 & 12.4 & 12.41 & 12.39
\end{tabular}
\end{table}

Before recording each dataset, the tank's turbidity level was measured using a portable turbidity meter kit (Sper Scientific). Between datasets, we used a wrist-mounted jetting tool powered by an electric pressure washer pump to stir up sediment from the bottom of the tank, which simulates the turbidity generated from a jetting excavation.

\section{Results}
\label{sec:results}
\begin{figure*}[h]
  \centering
\includegraphics[width=\linewidth]{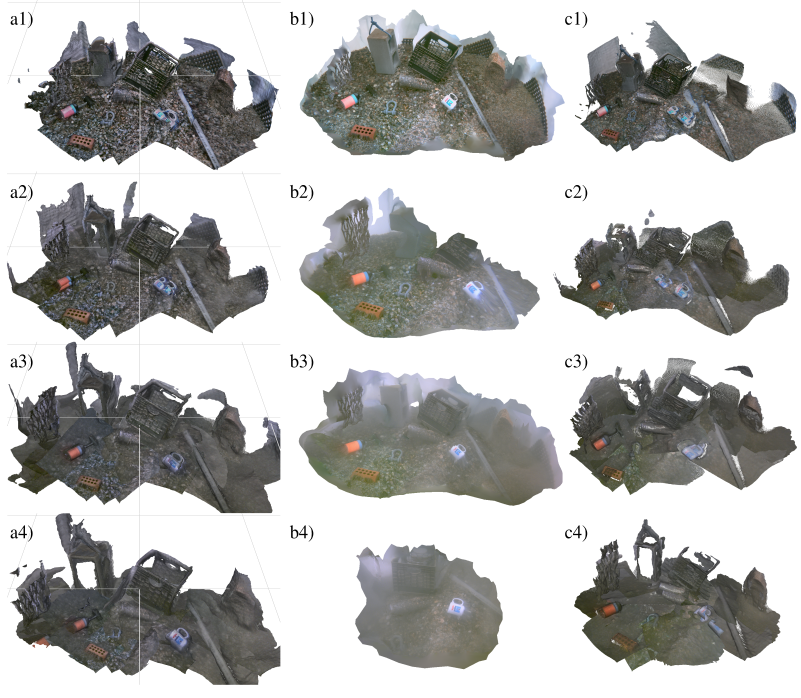}
  \caption{Optical 3D reconstruction results from (a) Sonar-MASt3R, (b) Metashape, and (c) MASt3R-SLAM using datasets A (1), C (2), E (3), and F (4) with turbidity values from $<$0.5 to 8 NTU. A 1-meter grid is included in Sonar-MASt3R's results for scale. No scale reference is provided for Metashape or MASt3R-SLAM since these methods do not produce metric-scale reconstructions.}
  \label{fig:compare-result}
\end{figure*}

To evaluate our method, we use the sonar data recorded using the ``sweep'' trajectory and the optical data recorded using the object-centric trajectory using the setup described in Section~\ref{sec:dataset}. Since existing opti-acoustic reconstruction methods cannot create dense reconstructions of complex workspaces in real-time, we benchmark our results using optical-only methods to assess reconstruction quality. Reconstruction results from using Sonar-MASt3R, MASt3R-SLAM, and Metashape on datasets A, C, E, and F (with turbidity values ranging from $<$0.5 - 8 NTU) are presented in Figure~\ref{fig:compare-result}.  

\begin{table*}[]
\centering
\caption{Measured object sizes in Sonar-MASt3R results for each dataset, in meters.}
\label{tab:scale}
\begin{tabular}{c|cccccccccc}
             & \begin{tabular}[c]{@{}c@{}}Cargo\\ Net\end{tabular} & \begin{tabular}[c]{@{}c@{}}Push\\ Core\end{tabular} & Brick & Shackle & \begin{tabular}[c]{@{}c@{}}Milk\\ Crate\end{tabular} & \begin{tabular}[c]{@{}c@{}}Cinder\\ Block\end{tabular} & Mug   & Pipe  & Stick & Rock  \\ \hline
A            & 0.204                                               & 0.241                                               & 0.154 & 0.090   & 0.261                                                & 0.330                                                  & 0.097 & 0.067 & 0.047 & 0.380 \\
B            & 0.180                                               & 0.248                                               & 0.156 & 0.092   & 0.263                                                & 0.309                                                  & 0.083 & 0.070 & 0.038 & 0.336 \\
C            & 0.201                                               & 0.273                                               & 0.171 & 0.106   & 0.262                                                & 0.336                                                  & 0.093 & 0.066 & 0.043 & 0.328 \\
D            & 0.201                                               & 0.275                                               & 0.166 & 0.111   & 0.291                                                & 0.343                                                  & 0.098 & 0.073 & 0.044 & 0.382 \\
E            & 0.251                                               & 0.258                                               & 0.147 & 0.107   & 0.309                                                & 0.379                                                  & 0.109 & 0.065 & 0.031 & 0.376 \\
F            & 0.291                                               & 0.247                                               & 0.138 & 0.083   & 0.346                                                & 0.465                                                  & 0.132 & 0.080 & 0.052 & 0.422 \\
G            & 0.257                                               & 0.327                                               & 0.146 & 0.102   & 0.293                                                & 0.310                                                  & 0.089 & 0.086 & ND    & NM    \\
H            & 0.211                                               & ND                                                  & 0.146 & 0.088   & 0.311                                                & ND                                                     & 0.102 & ND    & ND    & ND    \\ \hline
Mean         & 0.224                                               & 0.267                                               & 0.153 & 0.097   & 0.292                                                & 0.353                                                  & 0.100 & 0.072 & 0.042 & 0.371 \\
STD          & 0.038                                               & 0.030                                               & 0.011 & 0.010   & 0.030                                                & 0.055                                                  & 0.015 & 0.008 & 0.007 & 0.034 \\ \hline
Ground-Truth & 0.363                                               & 0.325                                               & 0.202 & 0.134   & 0.328                                                & 0.395                                                  & 0.118 & 0.097 & 0.038 & 0.479 \\
GT STD       & 0.010                                               & 0.002                                               & 0.001 & 0.001   & 0.001                                                & 0.001                                                  & 0.001 & 0.004 & 0.001 & 0.003 \\
\% Error     & 38                                                  & 18                                                  & 24    & 27      & 11                                                   & 11                                                     & 15    & 25    & 11    & 23   \\
\multicolumn{11}{c}{\begin{tabular}[c]{@{}c@{}}*NM = not measurable, ND = not detected, GT STD = Standard deviation for ground-truth measurements \end{tabular}}
\end{tabular}
\end{table*}

As illustrated by Figure~\ref{fig:compare-result}, Sonar-MASt3R's reconstruction results remain relatively unchanged at turbidity ranges from $<$0.5 - 8 NTU. Since the color of the cinder block roughly matches the color of the suspended sediment, its reconstruction is the first to degrade with higher turbidity values. Incorporating external pose information and the sonar-based scale corrections also improves the Sonar-MASt3R's geometric consistency. While the geometric consistency of Metashape’s reconstruction is higher than that of Sonar-MASt3R and MASt3R-SLAM (e.g., features such as the mug handle only appear once), this outcome is expected, as Metashape applies a global optimization over the entire dataset rather than operating incrementally. However, Metashape's performance in turbid conditions was the least consistent, yielding only partial reconstructions at 2.6 and 8 NTU. As an incremental optimization method that does not incorporate external pose estimates, MASt3R-SLAM exhibited the lowest geometric consistency, with several objects multiple times in the reconstruction (e.g., mug, pipe, milk crate, stick).

The MASt3R-SLAM and Sonar-MASt3R reconstructions are notably less hazy than the one produced by Metashape. This is likely due to the fact that these keyframe-based methods decide which frames to include in the 3D reconstruction based on the number of detected features, and thus it preferentially selects frames which were recorded at shorter stand-off distances. Meanwhile, Metashape blends the colors based on feature matches across all of the images, regardless of stand-off distance. The histogram equalization process used for color correction in Sonar-MASt3R produced slightly clearer colors than the uncorrected outputs of MASt3R-SLAM, although this improvement was less pronounced than the haze observed in Metashape’s reconstructions.

\begin{figure}[h]
  \centering
\includegraphics[width=0.9\linewidth]{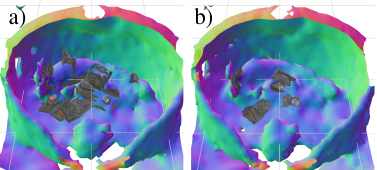}
  \caption{Sonar-MASt3R reconstruction results with dataset G (11.3 NTU) (a) and H (12.4 NTU) (b). The optical reconstruction is overlaid on the meshed acoustic reconstruction, which provides spatial context for the detected objects.}
  \label{fig:turbid-result}
\vspace{-20pt}
\end{figure}

At 11.3 NTU (dataset G), MASt3R-SLAM was unable to produce a 3D reconstruction since there were insufficient features for camera pose estimation. Metashape produced a partial reconstruction of this dataset which was missing the milk crate, cinder block, rock, and stick. At 12.4 NTU (dataset H), Metashape and MASt3R-SLAM were each unable to produce a reconstruction. In these datasets, Sonar-MASt3R adapted the use of optical and acoustic data based on the turbidity level, as illustrated in Figure~\ref{fig:turbid-result}. 
Sonar-MASt3R used dense optical features to add high-resolution features the acoustic reconstruction when visibility allowed, while maintaining robustness in turbid conditions by relying on sonar data and manipulator pose to maintain metric scale and preserve spatial relationships between objects. 



To quantitatively evaluate the scale of Sonar-MASt3R's reconstruction results, Table~\ref{tab:scale} reports object measurements using the reconstruction generated from each of the recorded datasets. The results highlight both the dimensional stability of the reconstructions despite the increasing turbidity and the ability to capture fine-scale structures (e.g., mug handle, shackle shape, milk crate pattern) smaller than a single voxel in the acoustic reconstruction. Although the object dimensions are self-consistent across the datasets, the objects' absolute scale in the reconstruction are consistently biased smaller than their actual size. 



We tested our implementation on an NVIDIA GeForce RTX 3070 GPU with an AMD Ryzen 9 5900HX CPU, where Sonar-MASt3R processed data at 3 FPS. Although MASt3R-SLAM reports a runtime of 15 FPS, we observed only 4 FPS when using their implementation on this hardware. This suggests that Sonar-MASt3R's processing speeds are likely comparable to MASt3R-SLAM, and that runtime differences can be attributed to GPU limitations rather than the methods themselves. For comparison, Metashape required approximately 25 minutes to reconstruct a dataset of $\sim$800 frames using this hardware.



\section{Discussion}
\label{sec:Discussion}


In an operational context, Sonar-MASt3R’s automated ability to adapt its reconstruction strategy based on visibility conditions makes it a promising approach for use in the field, particularly during intervention tasks that may generate transient turbidity plumes. As shown in Figure~\ref{fig:turbid-result}, although some objects were missing from the reconstruction in very turbid conditions, the acoustic reconstruction provided spatial context for coarse geometry between the detected objects. In the real-time visualization, we found that the projected data could be used to locate and identify objects even if they were not fully reconstructed or added to the keyframe map, which is similar to the approach used in~\cite{oasis}. Future work can add this image projection method as an automatic contingency step to enable object identification in conditions that are inadequate for dense optical 3D reconstruction.

By integrating external pose information and acoustic data, Sonar-MASt3R is capable of producing 3D reconstruction results with absolute spatial scaling in real time. Unlike previous opti-acoustic methods, it can reconstruct centimeter-scale objects with complex features in addition to the environment. The geometry of object features such as the brick holes, shackle loop, and mug handle are reconstructed, despite the coarse 5 cm acoustic voxel grid resolution.


For the acoustic 3D reconstruction, we re-implemented the OASIS method~\cite{oasis} to run on the GPU instead of the CPU, achieving processing rates above 100 FPS even with a 1 cm voxel grid resolution. However, we still used the lower voxel grid resolution of 5 cm in our experiments since the coarser resolution allowed each voxel to be sampled more frequently, which reduced false negatives. This also reduced GPU memory requirements, particularly during the acoustic reconstruction depth rendering process.
However, given that the scale of the objects were consistently smaller in the reconstruction, we hypothesize that the coarse grid resolution may have caused the acoustic depth render to bias the depth estimates closer, which would cause the objects to appear smaller.
Future work could examine the trade-offs of voxel grid resolution, and assess whether a higher-resolution depth rendering would improve dimensional accuracy of the reconstruction.


Sonar-MASt3R currently requires an external source of pose information, which works well for a fixed-base manipulator with joint angle sensors. However, future work on adapting this method to use a SLAM-based approach (e.g.,~\cite{singh2024opti}) could extend its applicability to free-floating platforms with greater pose uncertainty. Future work could also use a SLAM-based or another uncertainty-based approach to map dynamic environments, and track moving objects in the environment or changes made by the manipulator.


Compared to MASt3R-SLAM, we found that Sonar-MASt3R required using stricter pointmap confidence thresholds during keyframe selection and when adding edges to the factor graph. In MASt3R-SLAM, low threshold values were required to maintain tracking without interruptions, since a few untracked frames could cause the method to lose tracking and stall. In contrast, Sonar-MASt3R incorporates external pose and scale estimates from the manipulator and the sonar data, and thus mapping results are improved by discarding low-confidence frames with few features.



\section{Conclusion}
\label{sec:conclusion}
This paper presents Sonar-MASt3R, a novel opti-acoustic fusion method for real-time underwater 3D scene reconstruction in turbid conditions. By combining dense correspondences extracted from optical imagery using MASt3R with scale information derived from a rendered depth image using the acoustic reconstruction, this approach is able to generate metric-scale reconstructions in real time. Experimental evaluation across turbidity levels from $<$0.5 to $>$12 NTU using an eye-in-hand sensor configuration demonstrates improved robustness and reconstruction quality compared to baseline methods, which highlight the method's potential utility for robotic intervention operations in low visibility underwater environments. Future work will explore extending the approach to free-floating vehicles and dynamic environments.
\renewcommand{\bibfont}{\normalfont\small}
\printbibliography

\end{document}